\def\year{2019}\relax
\documentclass[letterpaper]{article} %DO NOT CHANGE THIS
\usepackage{aaai19}  %Required
\usepackage{times}  %Required
\usepackage{helvet}  %Required
\usepackage{courier}  %Required
\usepackage{url}  %Required
\usepackage{graphicx}  %Required
\usepackage{algorithm}
\usepackage{algorithmic}
\usepackage{amsmath}

\begin{document}
\title{Frame and Feature-Context Video Super-Resolution}
\author{Bo Yan\thanks{This work was supported by NSFC (Grant No.: 61772137; 61522202).}, Chuming Lin, Weimin Tan\\
School of Computer Science, Shanghai Key Laboratory of Intelligent Information Processing, Fudan University\\
\{byan, cmlin17, wmtan14\}@fudan.edu.cn\\
}
\maketitle
\begin{abstract}
For video super-resolution, current state-of-the-art approaches either process multiple low-resolution (LR) frames to produce each output high-resolution (HR) frame separately in a sliding window fashion or recurrently exploit the previously estimated HR frames to super-resolve the following frame. The main weaknesses of these approaches are: 1) separately generating each output frame may obtain high-quality HR estimates while resulting in unsatisfactory flickering artifacts, and 2) combining previously generated HR frames can produce temporally consistent results in the case of short information flow, but it will cause significant jitter and jagged artifacts because the previous super-resolving errors are constantly accumulated to the subsequent frames.

In this paper, we propose a fully end-to-end trainable frame and feature-context video super-resolution (\textbf{FFCVSR}) network that consists of two key sub-networks: local network and context network, where the first one explicitly utilizes a sequence of consecutive LR frames to generate local feature and local SR frame, and the other combines the outputs of local network and the previously estimated HR frames and features to super-resolve the subsequent frame. Our approach takes full advantage of the inter-frame information from multiple LR frames and the context information from previously predicted HR frames, producing temporally consistent high-quality results while maintaining real-time speed by directly reusing previous features and frames. Extensive evaluations and comparisons demonstrate that our approach produces state-of-the-art results on a standard benchmark dataset, with advantages in terms of accuracy, efficiency, and visual quality over the existing approaches.
\end{abstract}

The goal in image and video super-resolution (SR) is to reconstruct a high-resolution (HR) image or video from its down-sampled low-resolution (LR) version. Super-resolution approaches commonly serve as an important step for a variety of computer vision applications including image and video compression \cite{Li2017An,Kappeler2016Super}, medical imaging \cite{Yang2012Coupled}, object recognition \cite{Yang2018Long}, satellite imaging \cite{Demirel2011Discrete}, face recognition \cite{Gunturk2003Eigenface}, $etc$. To recover high-frequency details, single image super-resolution needs to fully exploit spatial statistics, while temporal correlations from multiple input frames are required to be exploited in order to improve reconstruction in the case of video super-resolution. Therefore, how to effectively exploit temporal redundancies becomes the key issue for video super-resolution.

\begin{figure}[t]
\centering
\includegraphics[width=7 cm]{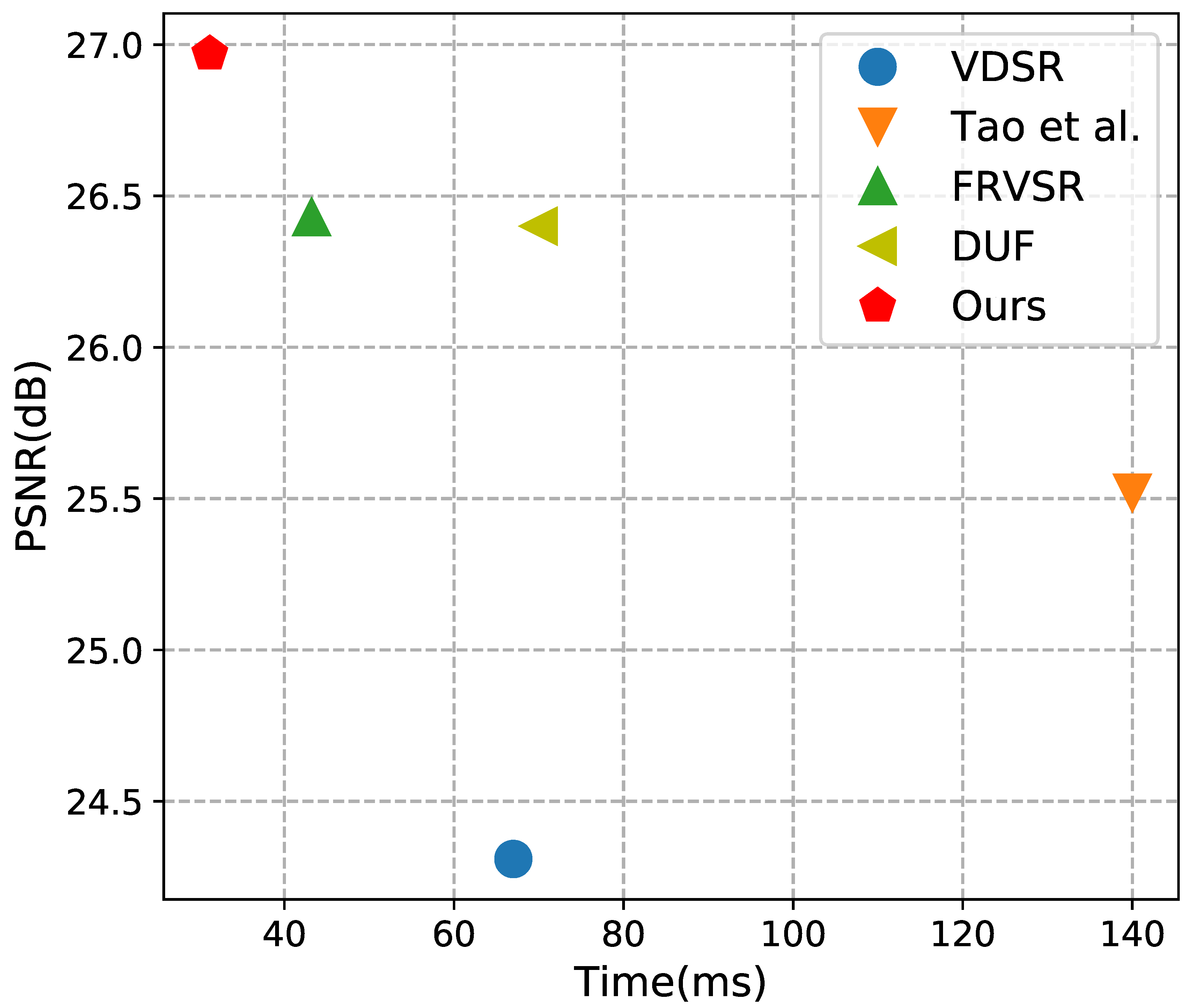}
\caption{The proposed approach consistently outperforms state-of-the-art video super-resolution methods in terms of reconstruction quality and efficiency (x4 SR on Vid4).}
\label{PSNRvsTime}
\end{figure}

Recent advances in video super-resolution are remarkable, benefiting mostly from the successful application of Deep Convolutional Neural Networks (DCNNs). However, there is still a large room for improvement over the DCNN based video super-resolution (SR) models that do not consider the super-resolution quality and temporal consistency simultaneously. The latest state-of-the-art approaches \cite{Dong2016Accelerating,VDSR2016cvpr,BayesSR2011cvpr,DESR2015iccv,VSRNet2016TCT,VESPCN2017cvpr,Tao2017iccv,DUF16L2018cvpr} formulate the task of video super-resolution as a great deal of separate multi-frame super-resolution subtasks. They exploit a sequence of consecutive LR frames to generate a single HR estimate, focusing on obtaining high-quality reconstruction results for each single frame. However, the way of separately generating each HR estimate results in temporally inconsistent frames, producing unsatisfactory flickering artifacts. In addition, such approaches have high computational cost because each input frame is processed several times.

To address the above issues, \cite{FRVSR2018cvpr} proposed a frame-recurrent video super-resolution model that recurrently exploits the previously estimated HR frames to super-resolve the subsequent frame. This approach is able to generate temporally consistent result because the current super-resolving frame will refer to those previously HR estimates. However, only referring to previously inferred HR frames will produce significant jitter and jagged artifacts because the previous super-resolving errors are constantly accumulated to the subsequent frames.

The above issues motivate us to develop a new approach for video super-resolution by introducing frame and feature as context to simultaneously improve the super-resolution quality and temporal consistency. Specifically, we design a novel end-to-end trainable video super-resolution framework that consists of two key sub-networks: local network and context network. The local network explicitly utilizes a batch of LR frames to generate local feature and local SR frame. Then, the context network combines the outputs of local network and the previously estimated HR frames and features to super-resolve the subsequent frame, which guides the network learning alignment between frames to maintain consistency. This framework offers three advantages:

\begin{itemize}
\item The inter-frame information from multiple LR frames can be effectively exploited by local network to generate high-quality SR frames (local SR frame) and reference features (local feature) that provides the context network higher-quality data to work with.
\item By utilizing the context information from previously predicted HR frames and features and the outputs of local network, our framework naturally encourages the video super-resolution model to generate temporally consistent results, making it to learn alignment between SR frames.
\item It has low computational cost due to its recurrent nature of using previous frames and features and no motion compensation block.
\end{itemize}

Benefiting from the property of combining context information from previous frames and features, the resulting architecture produces the most consistent results while containing finer details in each SR frame. Our model is fully convolutional and no other prior information such as optical flow estimation and motion compensation. To demonstrate the effectiveness of the proposed framework, we conduct ablation study for analyzing the importance of each component of our model. Besides, we compare our FFCVSR with several latest video super-resolution approaches and show that it produces state-of-the-art results on a standard benchmark dataset, with advantages in terms of accuracy, speed, and visual quality over the existing algorithms (see Fig. \ref{PSNRvsTime}). Furthermore, based on the characteristics of our framework, we propose a suppression-updating algorithm to effectively solve the problem of error accumulation of high frequency information. Finally, we also apply our trained model to real scenes to demonstrate its good abilities of generalization and practicability.

\section{Related Works}

Over the past decades, a large number of image and video super-resolution approaches have been developed, ranging from traditional image processing methods such as Bilinear and Bicubic interpolation to example-based frameworks \cite{Timofte2014Anchored,Jeong2015Multi,Xiong2013Example,Freedman2011TOG}, self-similarity methods \cite{Huang2015Single,Yang2010Exploiting}, and dictionary learning \cite{Perezpellitero2016PSyCo}. Some efforts have devoted to study different loss functions for high-quality resolution enhancement \cite{Sajjadi2017ICCV}. A complete survey of these approaches is beyond the scope of this work. Readers can refer to a recent survey \cite{Walha2016,Agustsson2017cvpr} on super-resolution approaches for details. Here, we focus on discussing recent video super-resolution approaches based on deep network.

Benefiting from the explosive development of convolutional neural network (CNN) in deep learning, CNN based approaches have refreshed the previous super-resolution state-of-the-art records. Since \cite{Dong2014Learning} uses a simple and shallow CNN to implement single super-resolution and achieves state-of-the-art results, following this fashion, numerous works have proposed various deep network architectures. Most of the existing CNN based video super-resolution approaches regard the task of video super-resolution as a large number of separate multi-frame super-resolution subtasks. They exploit a sequence of consecutive LR frames to generate a single HR estimate. \cite{VSRNet2016TCT} uses an optical flow method to warp video frames $LR_{t-1}$ and $LR_{t+1}$ onto the frame $LR_t$. Then, these three frames are combined to feed into a CNN model that outputs the HR frame $SR_t$. Similar to \cite{VSRNet2016TCT}, \cite{VESPCN2017cvpr} uses a trainable motion compensation network to replace the optical flow method in \cite{VSRNet2016TCT}. Following this fashion, Tao \textit{et al.} \cite{Tao2017iccv} propose a network comprising motion estimation, motion compensation, and detail fusion to process a batch of LR frames and output HR estimate.

%Following this fashion, Tao \textit{et al.} \cite{Tao2017iccv} propose a network comprising motion estimation, motion compensation, and detail fusion to process a batch of LR frames and output HR estimate. The motion estimation method is from \cite{VESPCN2017cvpr} and the motion compensation module is a subpixel motion compensation layer, which is similar to spatial transform network. The detail fusion module use a simple encoder-decoder style architecture with a Conv-LSTM in the center to produce HR estimate.

Different from the above mentioned approaches, \cite{FRVSR2018cvpr} proposes a frame recurrent video super-resolution (FRVSR) framework that combines the previous HR estimates to generate subsequent frame. This method warps the $SR_{t-1}$ frame onto the $SR_t$ based on the optical flow information estimated from $LR_{t-1}$ and $LR_{t}$. Then, it uses a trainable super-resolution network to fuse the warped $SR_{t-1}$ and $LR_{t}$, yielding the $SR_{t}$ frame. Therefore, there are two loss items in their loss function, the mean squared error between $SR_{t}$ and $HR_{t}$, and the warped $LR_{t-1}$ and $LR_{t}$. The FRVSR has advantage of producing temporally consistent results in the case of short information flow, but it will cause jitter and jagged artifacts because the previous super-resolving errors are constantly accumulated to the subsequent frames.

Though significant progress have been achieved by these studies in recent years, there is still a large room for improvement over the CNN based video super-resolution approaches that do not consider the super-resolution quality and temporal consistency simultaneously.

\begin{figure}[t]
\centering
\includegraphics[width=7 cm]{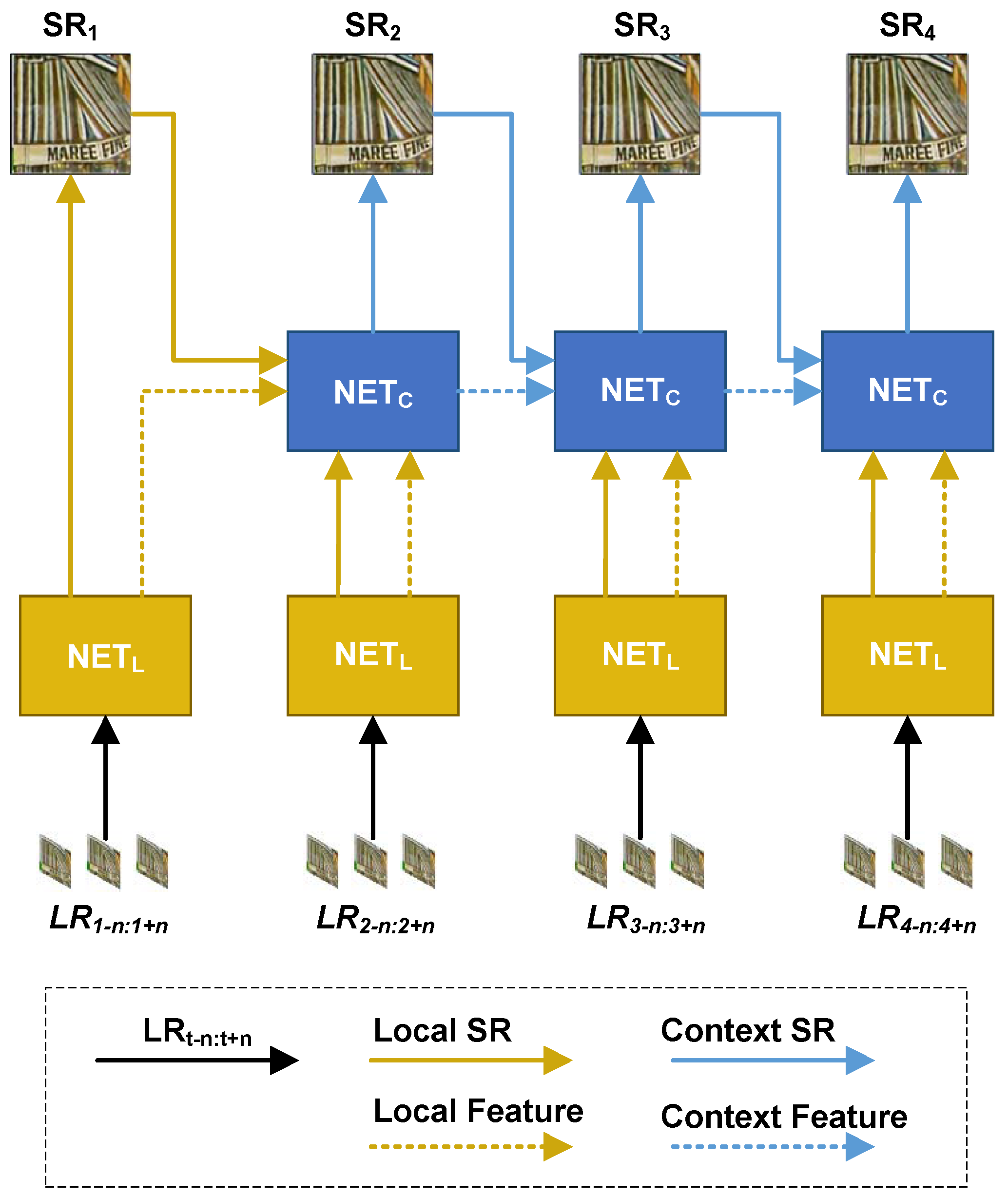}
\caption{Overview of the proposed FFCVSR framework. It consists of two trainable components: local network $NET_L$ (shown in yellow) and context network $NET_C$ (shown in blue). The $NET_L$ produces local frame $SR^{Local}_t$ and local feature $F^{Local}_t$ by processing a sequence of LR frames. Then, the $NET_C$ outputs the super-resolved result $SR_t$ and an additional output $F_t$. During training, the loss is applied on the output of $NET_L$ and $NET_C$, and back-propagated through both $NET_L$ and $NET_C$ for jointly training them.}
\label{proposedFramwork}
\end{figure}

\section{Method}

To overcome the aforementioned problems, we propose a frame and feature-context video super-resolution (\textbf{FFCVSR}) approach to reasonably combine both previous frames and features for accurate and fast video super-resolution. We will dedicate to state the proposed approach in detail in following subsections.

\subsection{FFCVSR Framework}

\textbf{1. Overview of the Proposed FFCVSR Framework:} To better understand our FFCVSR, we start out with the introduction of FFCVSR architecture, as illustrated in Fig. \ref{proposedFramwork}. It consists of two trainable components: local network $NET_L$ (shown in yellow) and context network $NET_C$ (shown in blue). Given a sequence of LR frames, the local network $NET_L$ outputs local frame $SR^{Local}_t$ and local feature $F^{Local}_t$ by exploiting inherent inter-frame information in the form of local correlations, helping the following context network $NET_C$ to recover lost high-frequency details. Considering the super-resolved results should maintain temporal consistency, the context network $NET_C$ not only exploits the local frame $SR^{Local}_t$ and previous SR frame $SR_t$ but also combines the local feature $F^{Local}_t$ and previous SR feature $F_t$, yielding visually pleasing and temporally consistent results. We will investigate the importance of each component by performing ablation study in the experiment, providing several insights for further designing better video super-resolution approach. Note that our FFCVSR framework has no motion compensation module commonly used in previous methods, which has additional advantage of reducing the computational cost. This processing flow is summarized in Algorithms \ref{alg1}.

\begin{algorithm}
\caption{Frame and Feature-Context Video Super-Resolution}
\label{alg1}
\begin{algorithmic}
\REQUIRE A sequence of consecutive LR frames, $LR_{t-n:t+n}$. \\
         ~~~~~~~~~~T is the updating step. T = 50 in our experiment.\\
\ENSURE Estimated high-resolution frame, $SR_t$.
\FOR{$t = 1 \to VideoLen$}
\IF{t == 1}
\STATE $SR_t \gets SR^{Local}_t$
\STATE $F_t \gets F^{Local}_t$
\ELSE
\STATE $(SR_t, F_t) \gets NET_C(SR_{t-1}, F_{t-1}, SR^{Local}_t, F^{Local}_t)$
\ENDIF
\ENDFOR
\STATE $\%$ Suppression updating algorithm
\IF{t mod T == 0}
\STATE $ SR_{t-1} \gets SR^{Local}_t$
\STATE $ F_{t-1} \gets F^{Local}_t$
\ELSE
\STATE $ SR_{t-1} \gets SR_t$
\STATE $ F_{t-1} \gets F_t$
\ENDIF
\end{algorithmic}
\end{algorithm}

\begin{figure*}[t]
\centering
\includegraphics[width=17.5 cm]{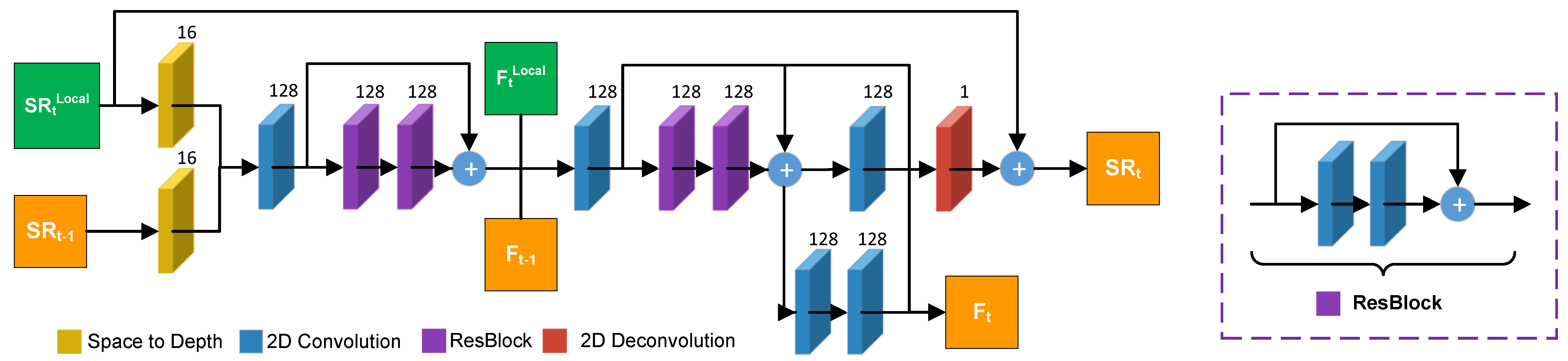}
\caption{Architecture of the proposed $NET_C$. It produces the HR frame $SR_t$ and feature $F_t$ by exploiting the context information from previously predicted HR frames and features ($i.e.$, $SR_{t-1}, F_{t-1}$) and the outputs of $NET_L$ ($i.e.$, $SR^{Local}_t, F^{Local}_t$).}
\label{figureContextNetwork}
\end{figure*}

\begin{figure}[t]
\centering
\includegraphics[width=8.5 cm]{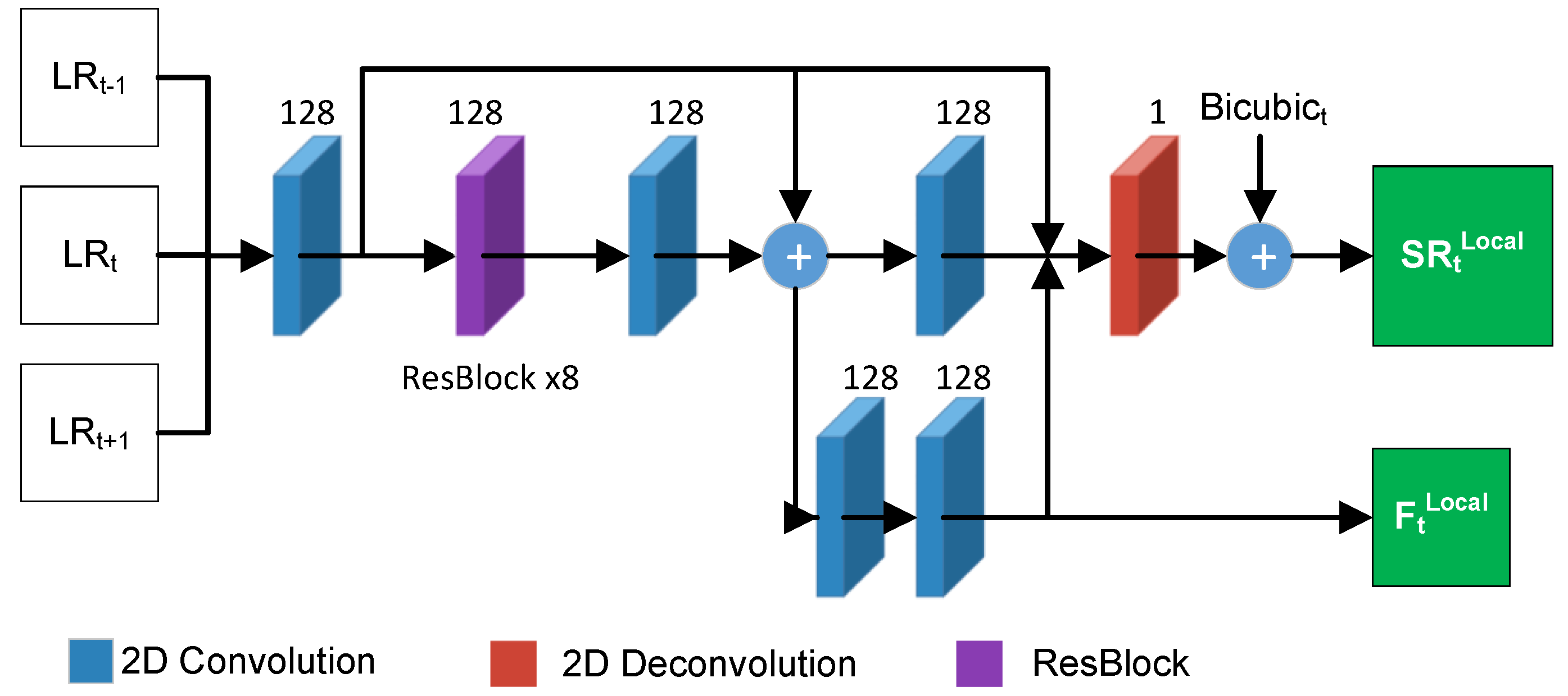}
\caption{Architecture of the proposed local network $NET_L$. It processes a sequence of LR frames to output local frame $SR^{Local}_t$ and local feature $F^{Local}_t$.}
\label{figureLocalNetwork}
\end{figure}

\textbf{2. Architecture of Local Network:} The proposed local network $NET_L$ is shown in Fig. \ref{figureLocalNetwork}. It exploits inherent inter-frame information in the form of local correlations and outputs local frame and feature by processing a sequence of LR frames. For demonstration convenience, we only show three consecutive LR frames including current frame that needs to be super-resolved. Our simple $NET_L$ consists of 5 convolutions (kernel size=$3\times3$, stride=1), 1 deconvolution (kernel size=$8\times8$, stride=4), and 8 ResBlocks \cite{Lim2017Enhanced}. The ResBlock in purple (shown on the right side of Fig. \ref{figureContextNetwork}) contains two convolutions with skip connection. We use the sum of the deconvolution result and the Bicubic interpolation result of $LR_t$ as the output $SR^{Local}_t$. The output $F^{Local}_t$ is produced by adding a new side output with two convolution operations. Let $T = {(LR_t, HR_t), t=1,\dotsc, N}$ denotes the training data set, where $LR_t$ is the input LR frame and $HR_t$ denotes the corresponding ground truth high-resolution frame. We use $W_L$ to denote the collection of all network layer parameters in $NET_L$. Thus, the local frame and feature can be given by:

\begin{equation}
  SR^{Local}_t, F^{Local}_t = NET_L(LR_{t-1}, LR_t, LR_{t+1}; W_L).
  \label{equationLocalNetwork}
\end{equation}

\textbf{3. Architecture of Context Network:} The proposed context network $NET_C$ is shown in Fig. \ref{figureContextNetwork}. It produces the HR estimate $SR_t$ and feature $F_t$ by exploiting the context information from previously predicted HR frames and features ($SR_{t-1}, F_{t-1}$) and the outputs of local network ($SR^{Local}_t, F^{Local}_t$), where the context information means that generating HR estimate will refer to previous HR frames and features to maintain temporal consistency. Our $NET_C$ consists of 5 convolutions (kernel size=$3\times3$, stride=1), 1 deconvolution (kernel size=$8\times8$, stride=4), 4 ResBlocks, and 2 space-to-depth transformations (shown in yellow) \cite{FRVSR2018cvpr}. Here, we use space-to-depth transformation to reduce the computational cost. We use the sum of the deconvolution result and local frame $SR^{Local}_t$ as the final output $SR_t$. We also provide another output of feature $F_t$ for super-resolving subsequent frame by adding a new side output with two convolution operations. We use $W_C$ to denote the collection of all network layer parameters in $NET_C$. Thus, the estimated HR frame and feature can be given by:

\begin{equation}
  SR_t, F_t = NET_C(SR_{t-1}, F_{t-1}, SR^{Local}_t, F^{Local}_t; W_C).
  \label{equationContextNetwork}
\end{equation}

\subsection{Loss Function}

The proposed local network $NET_L$ and context network $NET_C$ in our FFCVSR framework are seamlessly combined and jointly trained with the loss function defined as:

\begin{equation}
  Loss(W_L, W_C) = \left \| SR^{Local}_t - HR_t \right \|^2_2 + \left \| SR_t - HR_t \right \|^2_2.
  \label{LossFunction}
\end{equation}

The loss is applied on the output of $NET_L$ and $NET_C$, and back-propagated through both $NET_L$ and $NET_C$. Note that there is no need for defining additional loss function to constrain the output of features $F^{Local}_t$ and $F_t$, because as training progresses, both of them gradually provide high-quality data required by the $NET_C$ network.

\subsection{Suppression Updating Algorithm}

There is a key observation that the super-resolved video has significant jitter and jagged artifacts when using the previously inferred HR frames as reference information to generate subsequent frame, because the previous super-resolving errors are constantly accumulated to the subsequent frames. The Fig. \ref{SuppressingFigure} provides intuitive image examples for showing this observation. To overcome this problem, based on the characteristics of our FFCVSR framework, we propose a simple suppression-updating algorithm to effectively solve the problem of error accumulation of high frequency information. Specifically, we replace the $SR_{t-1}$ and $F_{t-1}$ outputted by $NET_C$ with $SR^{Local}_t$ and $F^{Local}_t$ outputted by $NET_L$ at each interval of $T$ frames, respectively (see also Algorithms \ref{alg1}), because after several iterations, the outputs of $NET_C$ have accumulated a considerable amount of super-resolving error while the outputs of $NET_L$ still maintain accurate information from current LR frame without introducing accumulative error from previous SR frames. In the experiment, we observe that $T = 50$ can produce favorable results.

\subsection{Training and Inference}

Our training dataset consists of 2 high-resolution videos (4k, 60fps): \textit{Venice} and \textit{Myanmar} downloaded from harmonic\footnote{https://www.harmonicinc.com/free-4k-demo-footage/}. The lengths of these two videos are 1,077 seconds and 527 seconds, respectively. We select them as training set because they contain more than 140 different scenes including human, natural scene, building, traffic, $etc$. To produce HR videos, we firstly downscale the original videos by factors of 4 ($960 \times 540$), 6 ($640 \times 360$), 8 ($480 \times 270$), 12 ($320 \times 180$), and 16 ($240 \times 135$) to obtain the high-resolution ground truth with a variety of receptive fields. Then, we extract patches of size $128 \times 128$ to produce the HR videos. To produce the input LR videos, we downsample them to the original 1/4 size using bilinear interpolation.

During training, we extract clips of 10 consecutive frames from the videos. We avoid the clips containing keyframes that have large scene changes. The extracted LR patches are randomly flipped horizontally and vertically for data augmentation. Besides, the order of sequences is also randomly reversed. We employ the brightness channel $y$ to train the proposed model. The parameters are updated with initial learning rate of $10^{-4}$ before 300K iteration steps and changed to $10^{-5}$ at the following 50K. The loss is minimized using Adam optimizer \cite{Kingma2014Adam} and back-propagated through both networks $NET_L$ and $NET_C$ as well as through time. After repeatedly minimizing the loss on the training data, the resulting network is capable of directly producing the full video frames, without needing any additional post-processing operations.

When super-resolving the first frame $SR_{t=1}$ in each clip, the local network $NET_L$ upsamples it at both training and testing time. At the same time, we regard the local frame and feature as the previously inferred frame and feature and feed them into the the context network $NET_C$ to produce $SR_t$ and $F_t$. This simple technique that reuses the outputs of $NET_L$ to deal with the first frame without prior information can encourage the network to exploit local information from LR frames during early training instead of only depending on the previously inferred HR estimates. Our architecture is fully end-to-end trainable and does not require pre-training sub-networks.

During inference, the trained model can process videos with arbitrary length and size due to the fully convolutional property of the networks. We can obtain the enhanced video by performing a single feedforward inference over frame by frame. In the following section, we report the reconstruction accuracy, efficiency, and visual quality of the model.

\section{Experiments and Analyses}  \label{ExperimentsResults}

In this section, we introduce compared methods and utilized dataset, and report the performance of our proposed approach. Firstly, we conduct ablation experiments to investigate the importance of each component of our approach, providing insight into how the performance of our FFCVSR varies with context information. Then, we compare our approach with current state-of-the-art methods on the standard \textbf{Vid4} benchmark dataset \cite{BayesSR2011cvpr} in terms of visual quality, objective metric, temporal consistency, and computational cost. Following \cite{VESPCN2017cvpr}, the evaluation metrics of Peak signal-to-noise ratio (\textbf{PSNR}) and structural similarity (\textbf{SSIM}) are computed on the brightness channel on the 4-video dataset Vid4. Thirdly, we detail the suppression-updating algorithm for depressing iteration error of high-frequency information. Finally, real-world examples are provided to verify the effectiveness of our approach. All experiments are carried out for 4x upscaling.

We conduct our experiments on a machine with an Intel i7-7700k CPU and an Nvidia GTX 1080Ti GPU. Our framework is implemented on the TensorFlow platform.

\begin{table}[t] \small %\scriptsize
\centering
\caption{Experimental result of ablation study. Average video PSNR of different architectures on Vid4.}
\begin{tabular}{p{2.4cm}p{0.85cm}p{0.7cm}p{0.7cm}p{0.7cm}p{0.7cm}}
\hline
Methods                                                                             & Calendar & City   & Walk   & Foliage & average \\ \hline
SISR                                                                                & 21.789   & 25.926 & 27.742 & 24.429  & 24.971         \\
\begin{tabular}[c]{@{}l@{}}only local network\end{tabular}        & 23.206   & 27.166 & 29.465 & 25.713  & 26.387         \\
w/o feature context                                                                 & 23.601   & 27.393 & 29.908 & 26.079  & 26.745         \\
\begin{tabular}[c]{@{}l@{}}w/o feature context+\\with optical flow\end{tabular} & 23.528   & 27.405 & 29.842 & 26.044  & 26.705         \\
\textbf{Full model}                                                                & \textbf{23.828}   & \textbf{27.564} & \textbf{30.172} & \textbf{26.296}  & \textbf{26.965} \\ \hline
\end{tabular}
\label{AblationStudy}
\end{table}

\subsection{Ablation Analysis}

Our architecture consists of two key components: local network $NET_L$ and context network $NET_C$. We experiment with different design options to illustrate the contribution of each component to the video super-resolution result in terms of objective metric and visual quality. To explore the performance of local network in the proposed architecture shown in Fig. \ref{proposedFramwork}, we remove the context network and use ``only local network'' to denote the resulting model, $i.e.$, blue arrows are removed in Fig. \ref{proposedFramwork}. Besides, we also explore the effectiveness of utilizing features including local feature (blue dashed arrow in Fig. \ref{proposedFramwork}) and context feature (yellow dashed arrow in Fig. \ref{proposedFramwork}) in the proposed framework. Thus, we remove the local feature and context feature from the architecture and use ``w/o feature context'' to denote it. Finally, we use ``Full model'' to denote our complete model including $NET_L$ and $NET_C$. Since optical flow method \cite{FRVSR2018cvpr} is widely employed in prior art, we incorporate it into our architecture to test whether it improves the recovering ability of our model. Here, we use ``w/o feature context + with optical flow'' to denote the model that removes feature information and introduces optical flow method. We also compare our model with a single image super-resolution (SISR) baseline, which is obtained by only feeding the current frame $LR_t$ into the local network.

\begin{table*}[t]
\centering
\caption{Quantitative comparison with state-of-the-art approaches. Values marked with a star are referenced from the corresponding publications. Obviously, our approach outperforms other methods in terms of PSNR, SSIM, and computational cost.}
\begin{tabular}{llllllllll}
\hline
Methods   & BayesSR* & DESR*   & VSRNet* & VESPCN* & VDSR*   & Tao et al.* & FRVSR   & DUF     & \textbf{Ours}  \\ \hline
% Reference & CVPR'11  & ICCV'15 & TCI'16  & CVPR'17 & CVPR'16 & ICCV'17     & CVPR'18 & CVPR'18 &       \\  \hline
x4 PSNR   & 24.42    & 23.50   & 22.81   & 25.35   & 24.31   & 25.52       & 26.43   & 26.40   & \textbf{26.97} \\
x4 SSIM   & 0.72     & 0.67    & 0.65    & 0.76    & 0.67    & 0.76        & 0.80    & 0.80    & \textbf{0.83}  \\
time (ms)  &  -       &   -     &  -      &   -      & 73.2    & 140         & 43.2    & 70      & \textbf{31.2}  \\ \hline
\end{tabular}
\label{ComparisonQuantitativeArts}
\end{table*}

\textbf{Quantitative Comparison:} The quantitative results are reported in Table \ref{AblationStudy}. The ``only local network'' model that exploits temporal information from input consecutive frames outperforms SISR baseline, which demonstrates that exploiting temporal redundancies is helpful to recover high-frequency details for video super-resolution. The ``w/o feature context'' model utilizing previously estimated HR frames further improves the performance of ``only local network''. This result implies that propagating information from previous HR frames to the following step helps the model to recover lost fine details. The complete model ``Full model'', simultaneously exploiting previously inferred frames and features, obtains the best results on all videos from Vid4. The result well demonstrates the effectiveness of introducing local and context features.

Compared with ``w/o feature context'', the ``w/o feature context + with optical flow'' method incorporating optical flow component leads to a slight decrease in PSNR. The possible reason is that the convolutional kernels are better at learning motion information from consecutive frames than optical flow method because of small motion of pixels in consecutive frames. We observe that directly using convolutional kernels to learn motion information instead of optical flow method not only improves the reconstruction accuracy of the model, but also decreases its computational cost. All the above experiments show that the proposed architecture for video super-resolution is reasonable and appropriate.

\begin{figure}[t]
\centering
\includegraphics[width=7 cm]{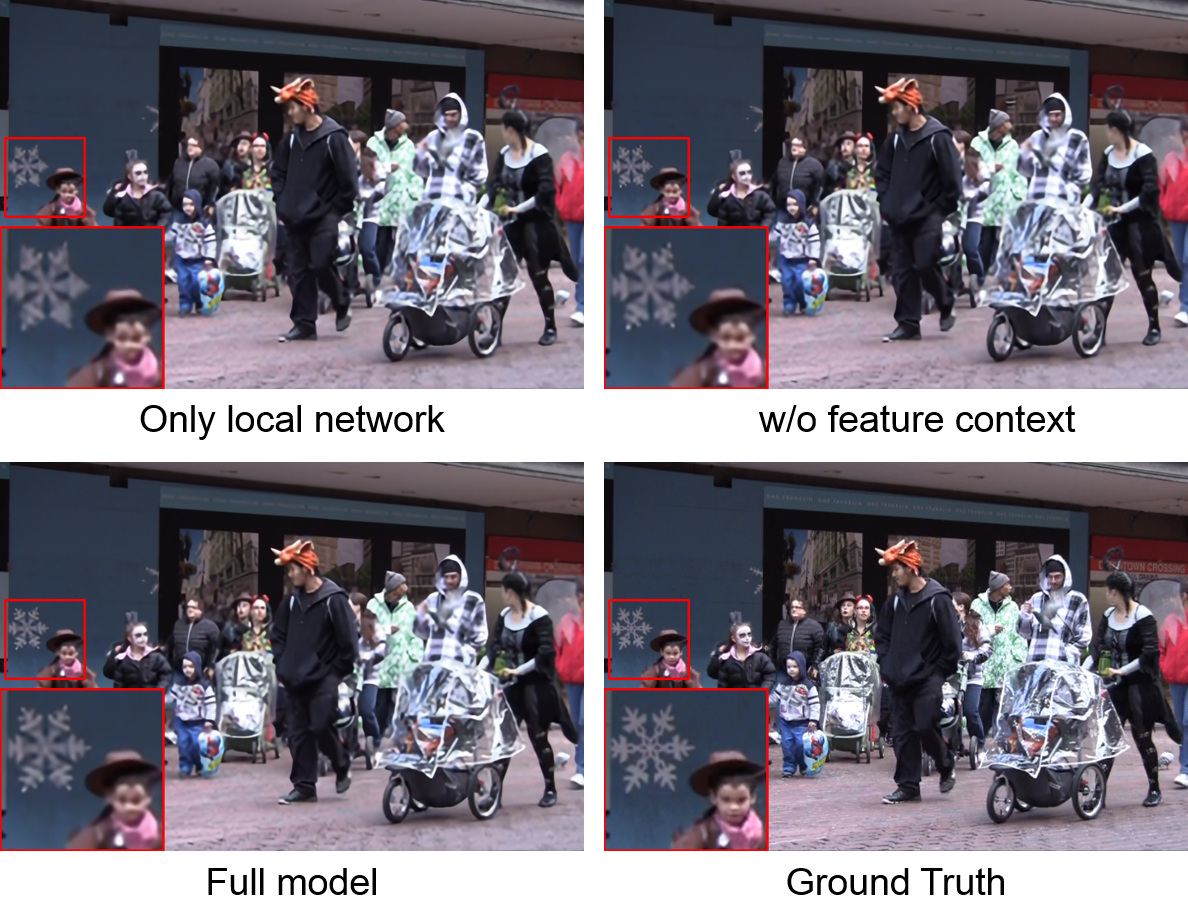}
\caption{A visual comparison of oblation study. The ``Full model'' produces the best result having fine details.}
\label{VisualAblationStudy}
\end{figure}

\textbf{Visual Comparison:} Figure \ref{VisualAblationStudy} shows a visual comparison of oblation study with different design options for our framework. We can observe that the ``Full model'' is capable of recovering finer details and generating visually satisfactory results. Compared with ``only local network'' and ``w/o feature context'' methods, the recovered result produced by ``Full model'' is both sharper and closer to the ground truth, as shown in the white snow in Fig. \ref{VisualAblationStudy}.

\begin{figure}[t]
\centering
\includegraphics[width=8.5 cm]{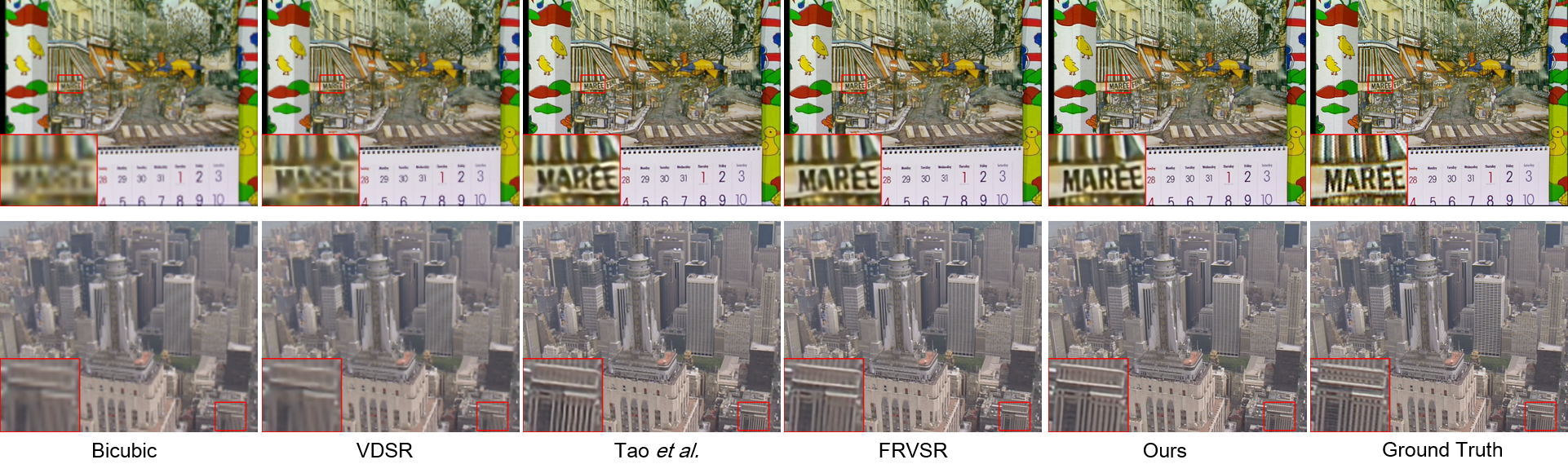}
\caption{Visual comparison with state-of-the-art approaches (x4 SR).}
\label{ComparisonVisualArts}
\end{figure}

\begin{figure}[t]
\centering
\includegraphics[width=8 cm]{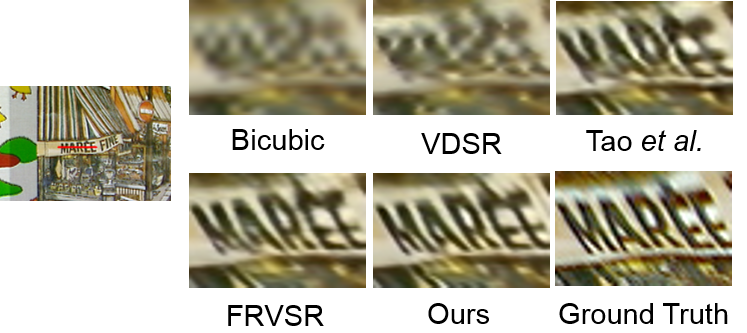}
\caption{Demonstration of temporal profiles for comparing temporal consistency of different approaches.}
\label{TemporalConsistency}
\end{figure}

\subsection{Comparison with prior art}

We compare the proposed approach with various state-of-the-art video super-resolution methods, including  \textbf{VDSR} \cite{VDSR2016cvpr}, \textbf{BayesSR} \cite{BayesSR2011cvpr}, \textbf{DESR} \cite{DESR2015iccv}, \textbf{VSRNet} \cite{VSRNet2016TCT}, \textbf{VESPCN} \cite{VESPCN2017cvpr}, \textbf{Tao \textit{et al.}} \cite{Tao2017iccv}, \textbf{FRVSR} \cite{FRVSR2018cvpr}, and \textbf{DUF} with 16 layer \cite{DUF16L2018cvpr} on the Vid4 benchmark dataset in terms of PSNR and SSIM. For all competing approaches except FRVSR and DUF, the PSNR and SSIM values are directly referenced from the corresponding publications by authors. Since FRVSR and DUF are the newest approaches, we implement them on the TensorFlow platform. For fair comparison, these two implements are trained and tested on the identical dataset used by our approach.

% there are no available implementations for them

\textbf{Quantitative Comparison:} Table \ref{ComparisonQuantitativeArts} reports the PSNR and SSIM produced by our approach and previous state-of-the-art methods on Vid4. It is obvious that our proposed approach substantially outperforms the current state-of-the-art methods by a large margin in terms of reconstruction accuracy and efficiency. Comparing with the current best results, our approach surpasses them by more than 0.5 dB in PSNR and 0.03 score in SSIM. This implies that our approach produces the most accurate result and our architecture is reasonable and appropriate for video super-resolution.

\textbf{Quality Comparison:} Fig. \ref{ComparisonVisualArts} demonstrates quality comparison of different approaches. From the close-up images, we observe that the proposed approach produces better structural detail than other competing methods. This result indicates that our strategy of exploiting previously inferred information in terms of frame and feature is essential such that the resultant SR images look much closer to the ground truth.

\subsection{Temporal Consistency}

To compare temporal consistency of different approaches, following \cite{VESPCN2017cvpr}, we use temporal profile to show the result on paper. Fig. \ref{TemporalConsistency} reports a temporal profile on the row highlighted by a red line across a number of frames. While \cite{Tao2017iccv} generates better results than VDSR method, it still contains considerable flickering artifacts due to separately estimating each output frame. By referring previous frames, FRVSR has improved a lot in the temporal consistency, but it has some blurs compared with the ground truth. In contrast, our approach produces the most temporal coherence result that looks much closer to the ground truth.

\begin{figure}[t]
\centering
\includegraphics[width=8 cm]{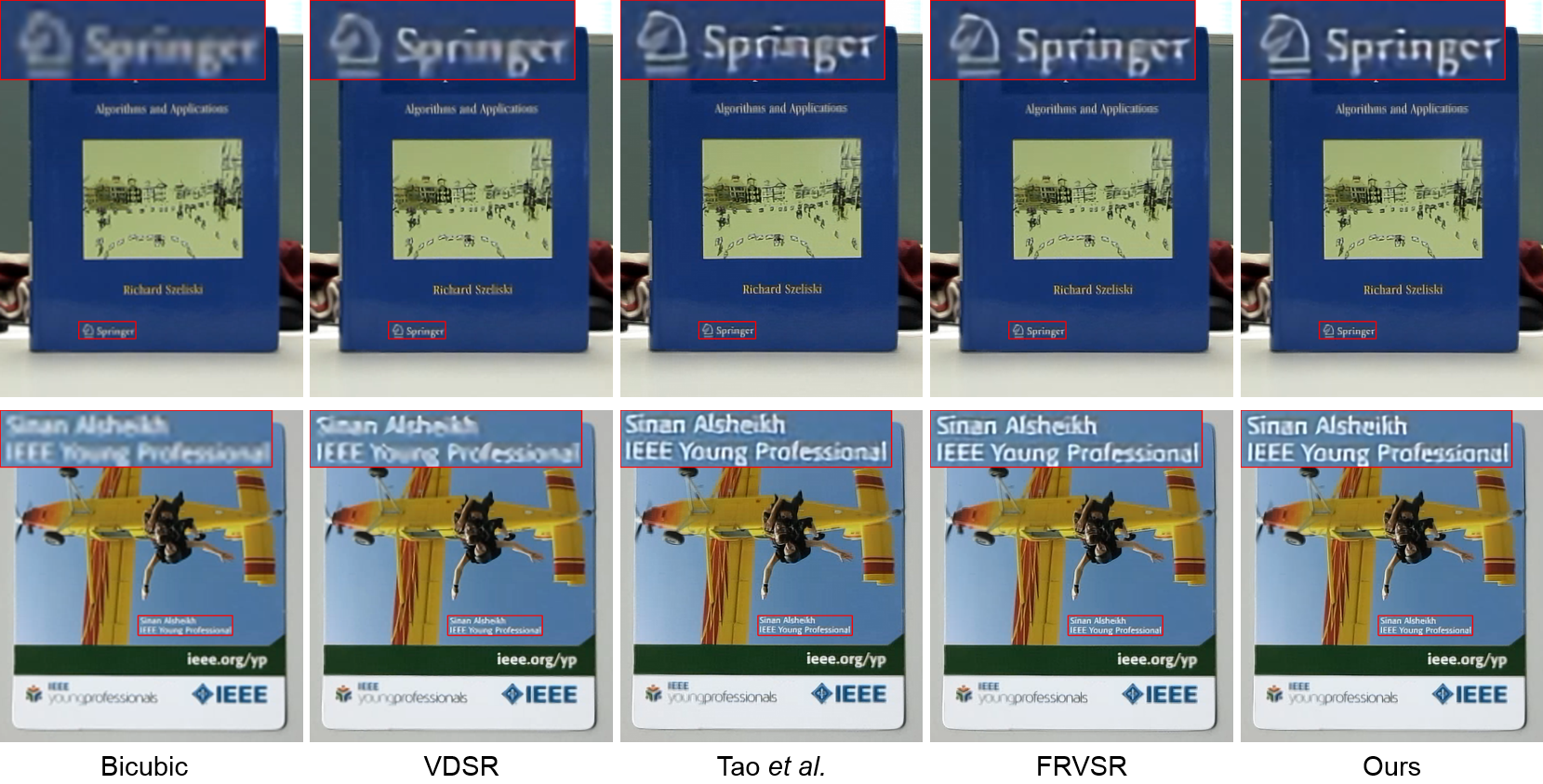}
\caption{Real-world examples to evaluate the practical ability of different approaches.}
\label{RealWorldExamples}
\end{figure}

\begin{figure}[t]
\centering
\includegraphics[width=6.1 cm]{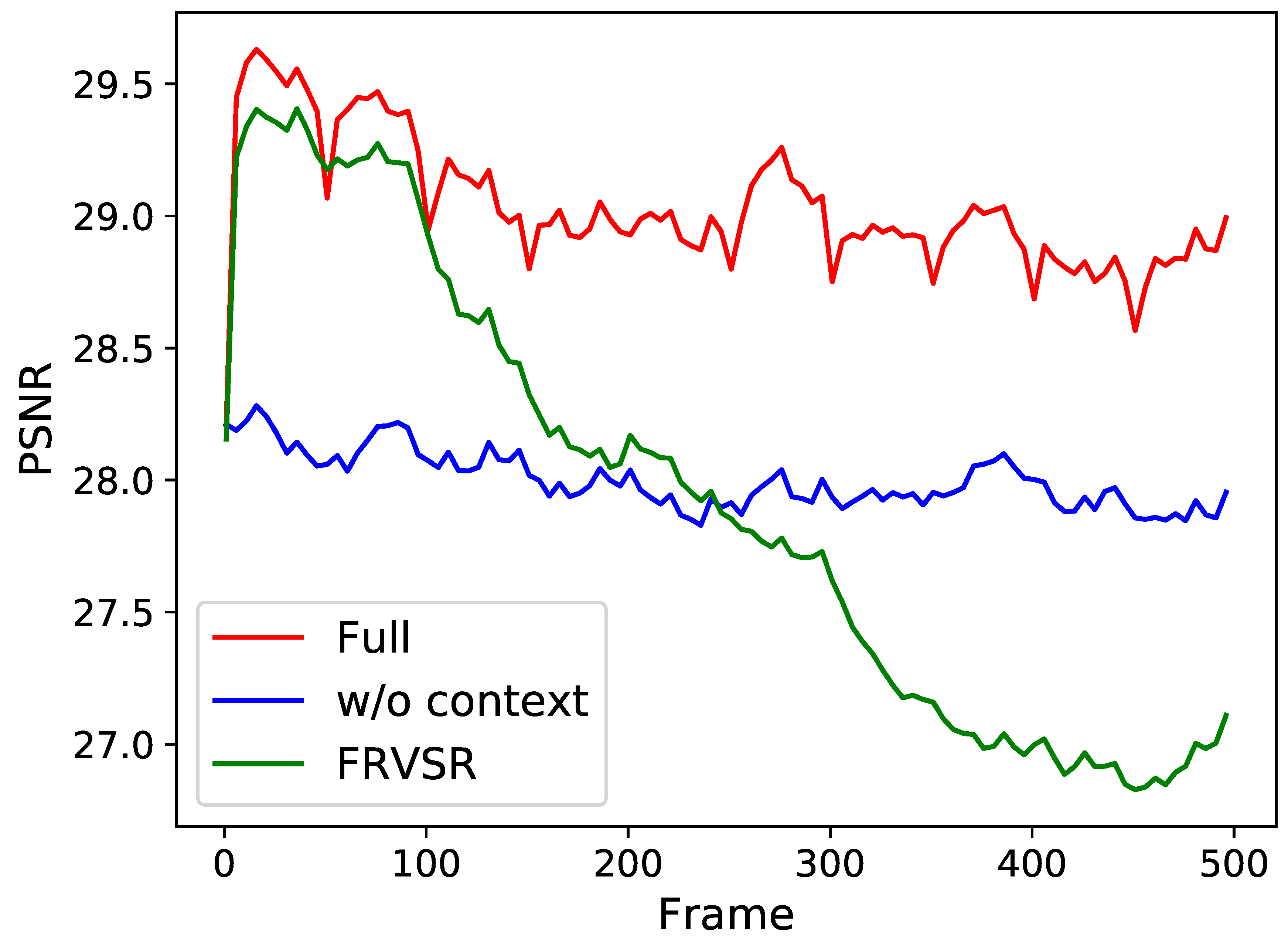}
\caption{Performance of FRVSR, our full model, and ``only local network'' on \textit{HongKong} as a function of the number of previous frames processed. Our suppression-updating algorithm can effectively depress iteration error of high-frequency information from previous frames processed.}
\label{SuppressingCurve}
\end{figure}

\begin{figure}[t]
\centering
\includegraphics[width=8.5 cm]{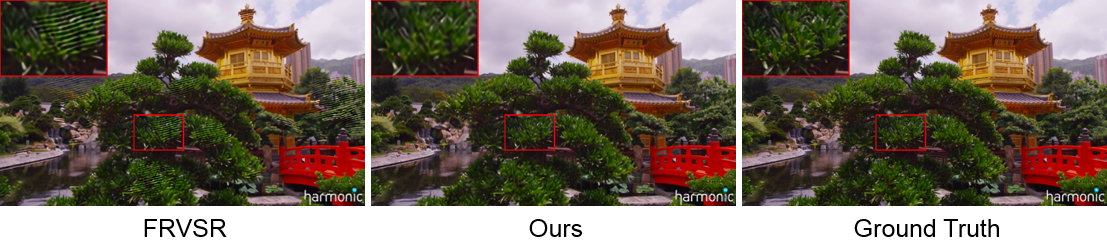}
\caption{Illustration of iteration error of high-frequency information.}
\label{SuppressingFigure}
\end{figure}

\subsection{Computational Efficiency}

Figure \ref{PSNRvsTime} and Table \ref{ComparisonQuantitativeArts} illustrate the comparison result of computational efficiency. Note that the running times of compared approaches including BayesSR \cite{BayesSR2011cvpr}, DESR \cite{DESR2015iccv}, VSRNet \cite{VSRNet2016TCT}, VESPCN \cite{VESPCN2017cvpr} are not listed in Table \ref{ComparisonQuantitativeArts}, because their running times are not stated in corresponding publications. The result clearly shows that our model is much more efficient than other approaches. It averagely takes 31.2 ms with our unoptimized TensorFlow implementation on an Nvidia GTX 1080Ti when running on Vid4 to generate a single HR image for 4x upsampling. Benefiting from directly taking advantage of previous features and frames, our approach is able to maintain real-time speed while producing high-quality temporal-coherency result.

\subsection{RealWorld Examples}

To evaluate the performance of our approach on real-world data, following \cite{VESPCN2017cvpr}, a visual comparison result is reported in Fig. \ref{RealWorldExamples}. From the close-up images, we observe that our approach is able to recover the fine details and remove the blur artifacts, even though the model is trained on a set of LR-HR frame pairs, where the LR frames are obtained by performing bicubic down-sampling.

\subsection{Suppressing Iteration Error of High-Frequency Information}

Because the previous super-resolving errors are constantly accumulated to the subsequent frames, the super-resolved video has significant jitter and jagged artifacts when using previously inferred HR frames. Fig. \ref{SuppressingCurve} illustrates the performance of FRVSR, our full model, and ``only local network'' (without context network) on \textit{HongKong}\footnote{https://www.harmonicinc.com/free-4k-demo-footage/} as a function of the number of previous frames processed. It shows that the reconstruction accuracy of FRVSR approach is high in the early stage and decreased slightly in the low range of information flow (less than 100 frames), but it decreases dramatically when the number of previous frames processed is over 100, even worse than our ``only local network''. In contrast, benefiting from the proposed suppression-updating algorithm, our full model and ``only local network'' are not affected by the number of previous frames processed and both achieve stable performance. Interestingly, the ``full model'' outperforms ``only local network'' method in all frames, which intuitively demonstrates the key contribution of the context network $NET_C$. Fig. \ref{SuppressingFigure} shows a visual comparison of iteration error of high-frequency information. Our approach effectively removes the unpleasing flickering artifacts existed in FRVSR method.

\section{Conclusion}

In this paper, we presented a frame and feature-context video super-resolution approach. Instead of only exploiting multiple LR frames to separately generate each output frame, we propose a fully end-to-end trainable framework consisting of local network and context network to simultaneously utilize previously inferred frames and features. Furthermore, based on the characteristics of our framework, we propose a suppression-updating algorithm to effectively solve the problem of error accumulation of high frequency information. Extensive experiments including ablation study demonstrate that our approach significantly advances the state-of-the-art on a standard benchmark dataset and is capable of efficiently producing high-quality temporal-consistency video resolution enhancement.

\bibliographystyle{aaai}
\bibliography{FFCVSR_ref}

\end{document}